\documentclass{article}

\usepackage[T1]{fontenc}
\usepackage[utf8]{inputenc}

\usepackage{ijcai17}

\usepackage[hidelinks]{hyperref}

\usepackage[ligature, inference]{semantic}
\usepackage{booktabs}
\usepackage{latexsym}

\usepackage{mathptmx}
\usepackage[scaled=.90]{helvet}
\usepackage{calrsfs}
\usepackage{courier}
\DeclareMathAlphabet{\pazocal}{OMS}{zplm}{m}{n}
\usepackage{microtype}
\usepackage{stmaryrd}
\usepackage{amssymb}
\usepackage{tabularx}
\usepackage[subtle]{savetrees}
\usepackage{paralist}

\usepackage[small]{caption}

\newenvironment{proofsketch}[1][Proof sketch]{\proof[#1]\mbox{}}{\endproof}

\usepackage{multicol}

\newlength\shlength
\newcommand\xshlongvec[2][0]{\setlength\shlength{#1pt}%
  \stackengine{-5.6pt}{$#2$}{\smash{$\kern\shlength%
    \stackengine{7.55pt}{$\mathchar"017E$}%
      {\rule{\widthof{$#2$}}{.57pt}\kern.4pt}{O}{r}{F}{F}{L}\kern-\shlength$}}%
      {O}{c}{F}{T}{S}}

\usepackage{mathtools}
\DeclarePairedDelimiter{\ceil}{\lceil}{\rceil}

\usepackage{tikz}
\usetikzlibrary{arrows,backgrounds,calc,chains,intersections,matrix,positioning,scopes}
\usetikzlibrary{shapes.misc}
\usepackage{amsthm}
\theoremstyle{definition}
\newtheorem{definition}{Definition}
\newtheorem{theorem}{Theorem}
\newtheorem{lemma}{Lemma}

\newtheorem{example}{Example}[section]

\usepackage{listings}
\usepackage{color}
\usepackage{xcolor}
\definecolor{labelcolor}{cmyk}{0.22,0.10,0.10,0.10}
\definecolor{listbackgroundcolor}{cmyk}{0.10,0.10,0.05,0.05}
\definecolor{listbackgroundcolorlight}{rgb}{0.91,0.92,0.94}
\definecolor{colorEntityBack}{rgb}{0.01,0.01,0.4}
\definecolor{colorPolicyBack}{rgb}{0.91,0.94,0.94}
\definecolor{colorApproachBack}{rgb}{0.93,0.93,0.97}

\newcommand{\commit}{\ensuremath\msf{commitment}}

\newcommand{\toward}{\ensuremath\msf{to}}

\newcommand{\create}{\ensuremath\msf{create}}

\newcommand{\detach}{\ensuremath\msf{detach}}

\newcommand{\discharge}{\ensuremath\msf{discharge}}

\newcommand{\fsf}[1]{\textsf{\small{#1}}}

\newcommand{\fsl}{\textsl}

\newcommand{\msf}{\mathsf}
\DeclareMathAlphabet{\mathsl}{OT1}{ptm}{m}{sl}

\newcommand{\mname}[1]{\fsl{#1}}
\newcommand{\pname}[1]{\fsl{#1}}
\newcommand{\rname}[1]{\textsc{#1}}

\newcommand{\ul}{\ulcorner}
\newcommand{\ur}{\urcorner}

\newcommand{\inn}{\ul\msf{in}\ur}
\newcommand{\out}{\ul\msf{out}\ur}

\newcommand{\key}{\ul\msf{key}\ur}

\newcommand{\bexa}{\begin{example}\rm} 
\newcommand{\eexa}{\end{example}}

\usepackage{xspace}
\newcommand{\etal}{{et al.\@\xspace}}

\newcommand{\lra}{\mbox{$\longrightarrow$ }}

\usepackage{fixme}
\fxsetup{
	status=draft,
	author={Note},
	layout=inline, 
	theme=color
}

\colorlet{hgreen}{black!30!green}
\colorlet{hred}{black!10!red}
\colorlet{hgrey}{gray!10!white}

\FXRegisterAuthor{tck}{atck}{\color{blue}TK}
\FXRegisterAuthor{mps}{amps}{MPS}

\title{Tosca: Operationalizing Commitments Over Information Protocols}

\author{
Thomas C.\ King$^{1}$ \and
Ak{\i}n G\"{u}nay$^{1}$ \and
Amit K.\ Chopra$^{1}$ \and
Munindar P.\ Singh$^{2}$ \\ \\
$^{1}$Lancaster University, Lancaster, LA1 4WA, United Kingdom \\
$^{2}$North Carolina State University, Raleigh, NC 27695-8206, USA \\
\{t.c.king, a.gunay, amit.chopra\}@lancaster.ac.uk, singh@ncsu.edu
}
\begin{document}

\maketitle

\begin{abstract}

  The notion of \emph{commitment} is widely studied as a high-level
  abstraction for modeling multiagent interaction.  An important
  challenge is supporting flexible decentralized enactments of
  commitment specifications.  In this paper, we combine recent
  advances on specifying commitments and \emph{information protocols}.
  Specifically, we contribute Tosca, a technique for automatically
  synthesizing information protocols from commitment specifications.
  Our main result is that the synthesized protocols support
  \emph{commitment alignment}, which is the idea that agents must make
  compatible inferences about their commitments despite
  decentralization.

\end{abstract}

\pagestyle{empty}
\thispagestyle{empty}

\section{Introduction}
\label{sec:intro}
Commitments represent a high-level abstraction for modeling multiagent
interaction \cite{Singh1999}.  The main idea
behind commitment protocols is to specify the 
\emph{meanings} of messages in terms of commitments
\cite{Pitt2001,Yolum2002}.  For example, to
capture a purchase, one may specify that a \fsl{Quote} message means
creating a commitment from the seller to the buyer to deliver an item
in exchange for payment.  In addition to meanings, a commitment protocol
typically also specifies operational constraints such as message
ordering and occurrence.  Thus, for example, one would specify that
the \fsl{Quote} message cannot occur before the \fsl{Request For
  Quote} message from the buyer to the seller.  Intuitively, the
motivation behind operational constraints is to rule out causally
invalid protocol enactments.

A fundamental challenge in this line of work has been supporting
\emph{decentralized} enactments of commitment protocols, that is, in
\emph{shared nothing} settings where agents communicate
\emph{asynchronously}.  Specifically, the only way for one agent to
convey information to another is to send it a message.
Supporting decentralized enactments in such settings is nontrivial
because agents may observe messages in incompatible orders.  Specifically, decentralization may
lead to situations where agents deadlock (lack of \emph{liveness}),
observe inconsistent messages (lack of \emph{safety}), or come to
incompatible conclusions about commitments that hold between them
(lack of \emph{alignment}
\cite{Chopra2008,Chopra2009,Chopra2015a})---all
three properties being crucial to interoperability.

Tosca addresses the challenge of decentralized enactments.  It builds
upon the conceptual observation that commitment specification
and operational constraints are distinct concerns
\cite{Chopra2008,Baldoni2013a}.  For
simplicity and clarity, from here on, we reserve \emph{protocol}
to mean an operational protocol specifying messages and the 
operational constraints on
their ordering and occurrence.  Specifically, the question Tosca
answers is: how can we operationalize commitment specifications over
protocols such that liveness and safety are preserved, 
and alignment is guaranteed? Tosca's contribution is a method 
for automatically synthesizing the appropriate protocol.  

Tosca's conceptual contribution is bringing three technical strands on
interaction in multiagent systems together.  One, BSPL
\cite{Singh2011}, a declarative language for specifying
protocols.  BSPL protocols are known as \emph{information protocols}
because ordering and occurrence constraints fall out from more fundamental
causality and integrity constraints on information in messages.  A
BSPL protocol can be checked for liveness and safety
\cite{Singh2012}.  Two, Cupid \cite{Chopra2015}, a
declarative language for specifying commitments.  The semantics of
Cupid is in terms of commitment-oriented queries on a relational
database.  Thus we may imagine an agent that runs (for whatever
purpose) commitment-oriented queries on its local database.  Three,
research on alignment \cite{Chopra2015a} (C\&S, for
brevity), which is about mechanisms for ensuring that the parties to a
commitment (the debtor and creditor) always progress toward
states where they make mutually compatible local inferences about the
commitment.  Specifically, whenever the creditor infers a commitment
as active from the messages it has observed, the debtor must as well infer it as active from its own observations.

Architecturally, Tosca brings the three strands together in the
following manner.  Each agent's local database or \emph{state}
comprises the messages it would have sent or received following a BSPL
protocol.  Cupid enables inferring the states of the commitments an
agent is party to from this database.  However, because each agent
carries out this inference on its own local state, it may turn out that
agents are not aligned with respect to a commitment.  Tosca
gives a method for ensuring progress toward alignment.  Specifically,
given a BSPL protocol and a set of commitments defined over the
messages in the protocol, it gives a method for synthesizing a BSPL
protocol whose enactment guarantees progress toward alignment.
Furthermore, if the input protocol is live and safe, the synthesized
protocol is live and safe as well.  

Tosca goes beyond C\&S in two ways.  One, it addresses
alignment for a more expressive language that includes deadlines,
nested commitments, and a richer commitment lifecycle.  Two, whereas
C\&S give algorithms for alignment, thereby constraining the
implementation of agents, Tosca gives a purely interactive solution in
terms of a protocol whose enactment would guarantee alignment.

\section{Background}

We now overview BSPL and Cupid, 
where for clarity we use \emph{message} (as in BSPL)
and \emph{commitment} (as in Cupid) to mean instances, and
\emph{specification} to mean the respective specifications.

\subsection{BSPL}
BSPL is used to declaratively specify protocols without explicit control flow.  By
contrast, languages such as AUML \cite{Huget2004} and RASA
\cite{Miller2007} rely on explicitly specifying message ordering. 
Instead, BSPL protocols impose information causality
constraints on each message $m$: what information $m$'s emission creates and what
information the sending role must know before sending $m$.
Thus, an implicit message ordering is imposed based solely on a protocol's explicit and declarative information causality specification.

Listing~\ref{list:simple-commit-input-protocol} demonstrates BSPL via the \mname{Ordering} protocol.
From here on, we describe such a protocol as 
an ``input protocol'' for Tosca, because it provides general message schemas 
for taking communicative actions (instantiating
messages) and it is distinguished from a synthesized protocol for aligning
a commitment.

The protocol \mname{Ordering} has the roles
\pname{M} (merchant), \pname{C} (customer), and \pname{S} (shipper); 
and the parameters \pname{oID} (order identifier), \pname{item},
\pname{price}, \pname{pID} (pay identifier), \pname{rID} (request
identifier), and \pname{sID} (ship identifier).

A \emph{complete} enactment of \mname{Ordering} comprises a
tuple of bindings for all of its parameters.
All parameters are adorned $\out$ for the protocol
as a whole, meaning that their values are bound by enacting the
protocol. Parameter \pname{oID} is annotated as a \fsf{key}
for the other parameters. This means each \pname{oID}
binding corresponds to a distinct tuple of bindings
for non key parameters and thus identifies
\mname{Ordering}'s enactment.
For example, it is not possible for the merchant to send two quotes with key
binding \pname{oID = 1} and different
non key parameter bindings.

\pname{Ordering} declares
four message schemas (their placement is irrelevant). By convention,
any key parameter of the protocol is a key parameter for any message
in which it appears.
The message schema \mname{quote} on Line~\ref{list:simple-commit-input-protocol:quote}
is from the merchant to the customer. It has three parameters, whose
values are bound by sending a quote due to being
adorned $\out$, namely \pname{oID}, \pname{item}, and
\pname{price}.

The message schema \mname{pay} on 
Line~\ref{list:simple-commit-input-protocol:pay} is from
the customer to the merchant. It comprises one
parameter adorned $\inn$, namely \pname{oID}, 
which means that its value binding must be known via
message emission or reception before a pay message is sent from 
the customer to the merchant. For example, 
the customer cannot send a \mname{pay} message with $\inn$
parameter binding \pname{oID = 1} before receiving a \mname{quote}
message with the same binding.
Hence, the customer can only send a \mname{pay} message after 
receiving a \mname{quote} from the merchant with the same key value,
based on the information (parameter) causality constraints.

Likewise, the message schema \mname{requestShip} on
Line~\ref{list:simple-commit-input-protocol:requestShip} 
is from the merchant to the
shipper and it has the $\inn$ parameter \pname{oID}.
Finally, \mname{ship}
on Line~\ref{list:simple-commit-input-protocol:ship}
has the \pname{oID} parameter adorned $\inn$. Since the shipper
can only know about \pname{oID}'s binding by receiving
a \mname{requestShip} message from the merchant, \mname{ship}
can only be sent after being requested.

\begin{lstlisting}[
  label={list:simple-commit-input-protocol},
  caption={A BSPL protocol for placing and fulfilling orders.},
  numbers=left,
  numbersep=1pt,
  xleftmargin=5pt,
  escapechar=|,
  numberstyle=\scriptsize]
Ordering{
 $\msf{roles}$ M, C, S // Merchant, Customer, Shipper
 $\msf{parameters}$ out oID key, out item, out price, out pID, out rID, out sID
 M $\mapsto$ C:quote[out oID, out item, out price] |\label{list:simple-commit-input-protocol:quote}|
 C $\mapsto$ M:pay[in oID, out pID] |\label{list:simple-commit-input-protocol:pay}|
 M $\mapsto$ S:requestShip[in oID, out rID] |\label{list:simple-commit-input-protocol:requestShip}|
 S $\mapsto$ C:ship[in oID, out sID] } |\label{list:simple-commit-input-protocol:ship}|
\end{lstlisting}

\subsection{Cupid}

Cupid is a language for specifying commitments over an event
database schema and inferring commitment
states based on an event database state.
In this paper, we only consider defining commitments over
protocol message schemas, where commitments are inferred over messages.

We demonstrate Cupid's basic ideas with an example
commitment in Listing~\ref{cupid:purchase} defined
on top of the message schemas of Listing~\ref{list:simple-commit-input-protocol}.

\begin{lstlisting}[label={cupid:purchase},caption={A
    specification in Cupid's surface syntax.}]
$\commit$ Purchase M $\toward$ C
 $\create$ quote
 $\detach$ pay[,quote + 10]
 $\discharge$ ship[,pay + 5]
\end{lstlisting}

A \emph{Purchase} commitment from \rname{M} (merchant) to
\rname{C} (customer) is \emph{created} when a quote is made. The
created commitment is uniquely
identified by \mname{quote}'s key (\pname{oID}). \emph{Purchase} is
\emph{detached} if a payment correlated to a quote occurs within ten time
points of the quote (\mname{pay} and \mname{quote} both have the same key, \pname{oID}). 
If the payment does not occur by the deadline, then the commitment is \emph{expired}
(failure to meet detach).  The commitment is \emph{discharged} if the
(correlated) shipment occurs within five time points of the payment;
if the shipment does not happen by the deadline,
the commitment is \emph{violated} (failure to meet discharge).
Cupid treats such lifecycle events as first-class events, meaning
that one commitment's lifecycle event may depend upon another's.

\subsection{Separation of Concerns}

Architecturally, Tosca uses BSPL to 
specify the \textit{operational}
layer interaction focusing on informational causality,
as a separate concern from the interaction \textit{requirements}
specified in Cupid. 
For example, we could change Listing~\ref{list:simple-commit-input-protocol}'s
\mname{requestShip} message schema to be causally dependent
on \mname{pay}'s identifier (\pname{pID}) before emission:

\begin{lstlisting}[
  numbers=left,
  numbersep=1pt,
  xleftmargin=5pt,
  escapechar=|,
  numberstyle=\scriptsize]
 M $\mapsto$ S: requestShip[in oID, in pID, out rID]
\end{lstlisting}

The modified information causality does not alter the fact that 
the \emph{Purchase} commitment continues to be discharged by a \mname{ship} 
message within five time points of the
\mname{pay} message. The modification does alter when these 
messages \textit{can} be sent. Conversely, changing the \emph{Purchase} commitment in
Listing~\ref{cupid:purchase}'s discharge
from \mname{ship} to another message would not
affect if and when \mname{ship} can be sent.

Tosca's separation of concerns supports modularity:
swapping out a protocol or modifying its information
causality does not affect commitment level requirements, only
when information (descriptively) \textit{can} be created and 
thus when commitments can be met; changing a commitment
does not affect if and when information can be created, 
only the (prescriptive) messaging \textit{requirements} between parties.

\section{Technical Motivation}

We now demonstrate commitment operationalization protocols,
which support realizing a commitment's lifecycle and progression
towards alignment via messaging. Specifically,
given a commitment defined over an input protocol that potentially causes misalignment,
we synthesize a commitment alignment protocol as output.

A commitment alignment protocol includes
the necessary message schemas for forwarding
messages to the creditor and debtor in order to
guarantee commitment alignment. Together, an input protocol
and multiple commitment alignment protocols are composed to form
a \emph{commitment operationalization protocol}.

\begin{figure}[t]
\centering
\begin{tikzpicture}[>=stealth]
\setlength{\fboxsep}{1pt}

\draw[thick] (-2.5,1.4) node[above] {M} -- (-2.5,-0.6);
\draw[thick] (-0.3,1.4) node[above] {C}  -- (-0.3,-0.6);
\draw [thick](-1.4,1.4) node[above] {S} -- (-1.4,-0.6);

\draw[->] (-2.5,1.2) -- (-0.3,0.9) node [midway, below, sloped] {\scriptsize{\colorbox{white}{$\text{quote}$}}};
\draw[thick, dashed, blue] (-2.9,1.1) node[left]{1} -- (0.1,1);
\draw[->] (-0.3,0.7) -- (-2.5,0.5)  node [midway, below, sloped] {\scriptsize{\colorbox{white}{$\text{pay}$}}};
\draw[thick, dashed, blue] (-2.9,0.6) node[left]{2} -- (0.1,0.6);
\draw[->] (-2.5,0.1) -- (-1.4,0)node [midway, below, sloped] {\scriptsize{\colorbox{white}{$\text{requestShip}$}}};
\draw[thick, dashed, blue] (-2.9,0.2) node[left]{3} -- (0.1,0.2);
\draw[->](-1.4,-0.2) -- (-0.3,-0.3)node [midway, below, sloped] {\scriptsize{\colorbox{white}{$\text{ship}$}}};
\draw[thick, dashed, blue] (-2.9,-0.4) node[left]{4} -- (0.1,-0.4);
\node at (-1.4,-1) {(\scriptsize{A})};

\draw[thick] (1.1,1.4) node[above] {M} -- (1.1,-0.6);
\draw[thick] (3.5,1.4) node[above] {C}  -- (3.5,-0.6);
\draw[thick](1.9,1.4) node[above] {S} -- (1.9,-0.6);
\draw[thick] (2.7,1.4) node[above] {E}  -- (2.7,-0.6);

\draw[->] (1.1,1.4) -- (3.5,1.3) node [midway, below, sloped] {\scriptsize{\colorbox{white}{$\text{quote}$}}};
\draw[->] (3.5,1.2) -- (2.7,1.1)  node [midway, below, sloped] {\scriptsize{\colorbox{white}{$\text{payEscrow}$}}};
\draw[thick, dashed, blue] (0.7,0.7) node[left]{5} -- (3.9,0.7);
\draw[->] (3.5,0.5) -- (1.1,0.4)  node [midway, below, sloped] {\scriptsize{\colorbox{white}{$\text{fwdCMPayEscrow}$}}};
\draw[thick, dashed, blue] (0.7,-0.1) node[left]{6} -- (3.9,-0.1);
\draw[->](1.1,-0.3) -- (1.9,-0.4)node [midway, below, sloped] {\scriptsize{\colorbox{white}{$\text{requestShip}$}}};
\draw[->](1.9,-0.5) -- (3.5,-0.6)node [midway, below, sloped] {\scriptsize{\colorbox{white}{$\text{ship}$}}};
\node at (2.4,-1) {(\scriptsize{B})};

\end{tikzpicture}
\caption{Protocol enactment for \textbf{M}erchant, 
\textbf{S}hipper, \textbf{E}scrow, and \textbf{C}ustomer roles.
}
\label{figSimpleAndForwarding}
\end{figure}
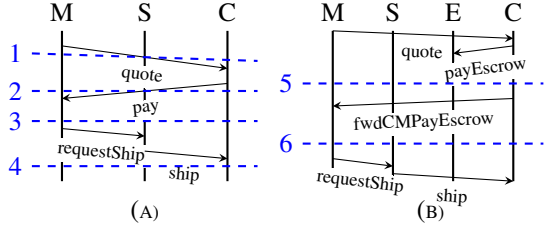

\subsection{Commitments Guaranteed Alignment}
\label{sec:SimpleCommitments}

The \emph{Purchase} commitment in Listing~\ref{cupid:purchase} is already alignable
for the create, detach, discharge, and expired lifecycle events by
the input protocol, \mname{Ordering}, 
in Listing~\ref{list:simple-commit-input-protocol}. In 
Figure~\ref{figSimpleAndForwarding}~(A), at time point~1
after the merchant sends the customer a \mname{quote}
but before the customer receives it,
the debtor (merchant) infers that the \emph{Purchase} commitment
is created. Hence the commitment is already aligned
(the debtor knows that they are committed) regardless
of what the creditor (customer) knows.

When the customer emits \mname{pay} before time point~2
they infer that the \emph{Purchase} commitment
is \textit{detached}. Hence, \emph{Purchase} becomes misaligned,
since the creditor (customer) has a stronger expectation of
the debtor (merchant) to discharge the commitment, which the debtor
does not know. The misalignment is rectified at time point~3, after
the merchant receives the \mname{pay} message.

Subsequently, after the merchant requests the shipper to ship,
the shipper sends a \mname{ship} message to the customer. 
At time point~4, after receiving
the \mname{ship} message, the creditor (customer) infers that the commitment
is discharged and hence does not have a stronger expectation
of the debtor (merchant) than what the debtor knows about (alignment).

Alignment for create, detach, discharge, and expired
lifecycle events is guaranteed, either because misalignment does
not occur (the creditor does not infer stronger
expectations of the debtor) or misalignment is rectified
via message reception. 
However, if ship is received after 
five time points of payment, then the creditor (customer)
infers violation whereas the debtor (merchant) cannot (permanent misalignment). 
Such misalignment requires message forwarding,
(e.g., notifying the debtor of ship), 
which we will cover in the next section.

\subsection{Commitments Requiring Forwarding}
\label{sec:CommitmentsRequiringForwarding}

Suppose an escrow service is used instead of direct payment from the
customer to the merchant. The \mname{EscrowOrdering} input protocol 
in Listing~\ref{list:input-thirdparty} 
and the \emph{EscrowPurchase} commitment in Listing~\ref{cupid:thirdparty} 
capture this situation.


\begin{lstlisting}[
  label={list:input-thirdparty},
  caption={An input protocol providing messaging for placing and carrying out 
  orders using an escrow service.},
  numbers=left,
  numbersep=1pt,
  xleftmargin=5pt,
  escapechar=|,
  numberstyle=\scriptsize
]
EscrowOrdering{
 $\msf{roles}$ E, M, C, S // Escrow, Merchant, Customer, Shipper
 $\msf{parameters}$ out oID key, out item, out price, out pID, out rID, out sID, out tID

 M $\mapsto$ C:quote[out oID, out item, out price]
 C $\mapsto$ E:payEscrow[in oID, out pID] |\label{list:input-thirdparty:payescrow}|
 M $\mapsto$ S:requestShip[in oID, out rID] |\label{list:input-thirdparty:requestShip}|
 S $\mapsto$ C:ship[in oID, out sID] |\label{list:input-thirdparty:ship}|
 E $\mapsto$ M:payTransfer[in oID, in pID, out tID] } |\label{list:input-thirdparty:transfer}|
\end{lstlisting}

\begin{lstlisting}[
  label={cupid:thirdparty},
  caption={A commitment to capture escrow payment.},
]
$\commit$ EscrowPurchase M $\toward$ C
 $\create$ quote
 $\detach$ payEscrow[, quote + 10]
 $\discharge$ ship[, payEscrow + 5]
\end{lstlisting}

In this scenario, we need to introduce a message that forwards
another message's occurrence to an otherwise ignorant party.
Listing~\ref{list:synthesized-thirdparty} shows a protocol that
introduces message forwarding in order to align
the \emph{EscrowPurchase} commitment (Listing~\ref{cupid:thirdparty})
for the input protocol, \mname{EscrowOrdering}
in Listing~\ref{list:input-thirdparty}. Specifically, by incorporating
the message schema, \mname{fwdCMPayEscrowID},
for \textit{forwarding} \mname{payEscrow} from the customer
to the merchant. Each forwarding message schema has a distinct name mapped 
to the message being forwarded.

To exemplify, in Figure~\ref{figSimpleAndForwarding}~(B) the customer sends
\mname{payEscrow} to the escrow.
At time point~5 we have misalignment,
because the customer (creditor) infers the
\emph{EscrowPurchase}'s detach and an expectation for the merchant 
to ship the goods. Yet the debtor (merchant) cannot
know the customer's expectation without notification. 
Misalignment is resolved at time point~6 once 
the customer forwards \mname{payEscrow} to the merchant 
via \mname{fwdCMPayEscrow}. Both roles know that the 
debtor (merchant) is expected to discharge the commitment (alignment).

\begin{lstlisting}[
  label={list:synthesized-thirdparty},
  caption={A protocol for aligning the \mname{EscrowPurchase} commitment in Listing~\ref{cupid:thirdparty}.},
  numbers=left,
  numbersep=1pt,
  xleftmargin=5pt,
  escapechar=|,
  numberstyle=\scriptsize
]
EscrowPurchaseAl{
 $\msf{roles}$ C, M
 $\msf{parameters}$ in oID key, in pID, out fwdCMPayEscrowID

 C $\mapsto$ M:fwdCMPayEscrow[in oID, in pID, 
                   out fwdCMPayEscrowID] } |\label{list:synthesized-thirdparty:fwd}|
\end{lstlisting}

\subsection{Nested Commitments}
\label{sec:NestedCommitments}

We now consider the case where one commitment's lifecycle event depends
upon another's (nesting). The \emph{EscrowTransfer} commitment given
in Listing~\ref{cupid:nested} is defined over the message schemas
from the input protocol \mname{EscrowOrdering} 
in Listing~\ref{list:input-thirdparty}. 
The escrow service is committed to the merchant to transfer the
customer's payment, once \emph{EscrowPurchase} is discharged.

\begin{lstlisting}[
  label={cupid:nested},
  caption={Escrow service's commitment to the merchant.}
]
$\commit$ EscrowTransfer E $\toward$ M
 $\create$ payEscrow
 $\detach$ discharged(EscrowPurchase)
 $\discharge$ payTransfer[, discharged(EscrowPurchase) + 5]
\end{lstlisting}

\emph{EscrowTransfer} is operationalized with the protocol 
in Listing~\ref{list:synthesized-nested}.
Focusing on the nested lifecycle event, 
the idea is that if \emph{EscrowTransfer}'s 
creditor (merchant) infers its detach, then so should the
debtor (escrow). \emph{EscrowTransfer}'s detach
is \emph{EscrowPurchase}'s discharge. Hence we ensure that whenever a
message contributes to \emph{EscrowPurchase}'s discharge it can be
forwarded to \emph{EscrowTransfer}'s debtor (escrow).

\begin{lstlisting}[
  label={list:synthesized-nested},
  caption={A protocol for aligning the \mname{EscrowTransfer} 
  commitment in Listing~\ref{cupid:nested}.},
  numbers=left,
  numbersep=1pt,
  xleftmargin=5pt,
  escapechar=|,
  numberstyle=\scriptsize
]
EscrowTransferAl{
  $\msf{roles}$ C, E, M //Customer, Escrow, Merchant
  $\msf{parameters}$ in oID key, in item, in price, in pID, in sID, out fwdMEQuoteID, out fwdCMPayEscrowID, out fwdSEShipID, out fwdMEShipID
  
  M $\mapsto$ E:fwdMEQuote[in oID, in item, in price, out fwdMEQuoteID]  |\label{list:synthesized-nested:fwdQuote}|
  C $\mapsto$ M:fwdCMPayEscrow[in oID, in pID, out fwdCMPayEscrowID] |\label{list:synthesized-nested:fwdPayEscrow}|
  S $\mapsto$ E:fwdSEShip[in oID, in sID, out fwdSEShipID] } |\label{list:synthesized-nested:fwdSEShipment}|
  M $\mapsto$ E:fwdMEShip[in oID, in sID, out fwdMEShipID] } |\label{list:synthesized-nested:fwdMEShipment}|
\end{lstlisting}

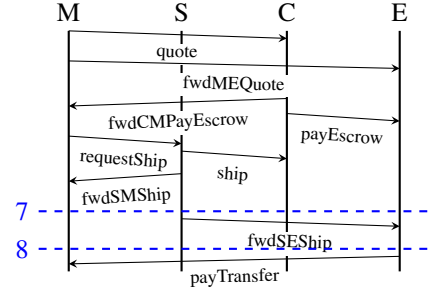
\begin{figure}[t]
\centering
\begin{tikzpicture}[>=stealth]
\setlength{\fboxsep}{1pt}

\draw[thick] (-1.7,-2.3) node[above] {M} -- (-1.7,-5.5);
\draw[thick] (1.2,-2.3) node[above] {C}  -- (1.2,-5.5);
\draw [thick](-0.2,-2.3) node[above] {S} -- (-0.2,-5.5);
\draw[thick] (2.7,-2.3) node[above] {E}  -- (2.7,-5.5);

\draw[->] (-1.7,-2.3) -- (1.2,-2.4) node [midway, below, sloped] {\scriptsize{\colorbox{white}{$\text{quote}$}}};

\draw[->] (-1.7,-2.7) -- (2.7,-2.8) node [midway, below, sloped] {\scriptsize{\colorbox{white}{$\text{fwdMEQuote}$}}};

\draw[->] (1.2,-3.4) -- (2.7,-3.5)  node [midway, below, sloped] {\scriptsize{\colorbox{white}{$\text{payEscrow}$}}};

\draw[->] (1.2,-3.2) -- (-1.7,-3.3)  node [midway, below, sloped] {\scriptsize{\colorbox{white}{$\text{fwdCMPayEscrow}$}}};
\draw[->](-1.7,-3.7) -- (-0.2,-3.8)node [midway, below, sloped] {\scriptsize{\colorbox{white}{$\text{requestShip}$}}};
\draw[->](-0.2,-3.9) -- (1.2,-4)node [midway, below, sloped] {\scriptsize{\colorbox{white}{$\text{ship}$}}};
\draw[->](-0.2,-4.2) -- (-1.7,-4.3)node [midway, below, sloped] {\scriptsize{\colorbox{white}{$\text{fwdSMShip}$}}};
\draw[thick, dashed, blue] (-2.1,-4.7) node[left]{7} -- (3.1,-4.7);
\draw[->](-0.2,-4.8) -- (2.7,-4.9)node [midway, below, sloped] {\scriptsize{\colorbox{white}{$\text{fwdSEShip}$}}};
\draw[thick, dashed, blue] (-2.1,-5.2) node[left]{8} -- (3.1,-5.2);

\draw[->](2.7,-5.3) -- (-1.7,-5.4)node [midway, below, sloped] {\scriptsize{$\text{payTransfer}$}};

\end{tikzpicture}
\caption{Protocol enactment for \textbf{M}erchant, 
\textbf{S}hipper, \textbf{E}scrow, and \textbf{C}ustomer roles where
each message shares key values.}
\label{figNestedInteraction}
\end{figure}

In Figure~\ref{figNestedInteraction} at time point~7, the
merchant infers \emph{EscrowPurchase}'s discharge and
consequently \emph{EscrowTransfer}'s detach.
Specifically, due to knowing about the messages contributing
to \emph{EscrowPurchase}'s discharge: \mname{quote}
(by sending it), \mname{payEscrow} (via the forwarding
message \mname{fwdCMPayEscrow}), and \mname{ship} (via
the forwarding message \mname{fwdSMShip}). 
Yet the debtor (escrow) neither 
infers \emph{EscrowPurchase}'s discharge
nor consequently \emph{EscrowTransfer}'s detach.
Hence, the creditor (merchant) expects the debtor (escrow)
to discharge \emph{EscrowTransfer}, which the debtor
is unaware of (misalignment).

Forwarding message schemas to the escrow support
rectifying the misalignment. \emph{EscrowPurchase}'s discharge is due to:
\mname{quote}, which can be forwarded to the escrow; \mname{payEscrow},
which the escrow receives (hence no forwarding is required)
and \mname{ship}, which can be forwarded to the escrow. 
In Figure~\ref{figSimpleAndForwarding} at time point 8,
after sending the forwarding messages, escrow knows about
\emph{EscrowPurchase}'s discharge and thus \emph{EscrowTransfer}'s detach
(alignment).

\subsection{Compositionality}

Commitment alignment protocols can be synthesized
independently and then composed.
In Listing~\ref{list:synthesized-compositional},
our preceding commitment alignment protocols
are subprotocols of an overall operationalization protocol.
If two commitment alignment protocols bind the same
$\out$ parameter, it is due to the same message being sent
and hence the binding is the same. For example
\pname{fwdCMPayEscrowID} is bound by
\mname{EscrowPurchaseAl} when a forward
message \mname{fwdCMPayEscrow} is sent
if and only if \pname{fwdCMPayEscrowID} is bound with
the same value by \mname{EscrowTransferAl} when the same
\mname{fwdCMPayEscrow} message is sent. Hence, independently
constructed alignment protocols are composed together without 
contradictory parameter bindings during enactment.

\begin{lstlisting}[
  label={list:synthesized-compositional},
  caption={An operationalization protocol composed
  from an input protocol and synthesized alignment protocols.},
  numbers=left,
  numbersep=1pt,
  xleftmargin=5pt,
  escapechar=|,
  numberstyle=\scriptsize
]
OperationalizationProtocol{
 $\msf{roles}$ M, C, E, S
 $\msf{parameters}$ out oID key, out item, out price, out pID, out sID, out rID, out tID, out fwdMEQuoteID, out fwdCMPayEscrowI, out fwdSEShipID, out fwdMEShipID
 
 EscrowOrdering(M, C, E, S, out oID, out item, out price, out pID, out sID, out rID, out tID)
 EscrowPurchaseAl(E, M, C, S, in oID, in pID, out fwdCMPayEscrowID)
 EscrowTransferAl(C, E, M, S, in oID, in item, in price, in pID, in sID, out fwdMEQuoteID, out fwdCMPayEscrowID, out fwdSEShipID, out fwdMEShipID) }
\end{lstlisting}

\subsection{Summary}

Tosca synthesizes the alignment protocol 
for a commitment and an input protocol. The alignment protocol
comprises forwarding message schemas, supporting participants
in aligning the commitment via messaging. 
Multiple commitment alignment 
protocols are composed together, without parameter interference, 
into an operationalization protocol for triggering commitment
lifecycles and supporting alignment via messaging.

\section{Synthesizing Protocols}

\subsection{Protocols}

We adopt BSPL's formal syntax from \cite{Singh2012}. We use the
following lists treated as sets: public
roles $\vec{x}$, private roles $\vec{y}$, public parameters $\vec{p}$,
$\key$ parameters $\vec{k} \subseteq \vec{p}$, $\inn$ parameters
$\vec{p}_I \subseteq \vec{p}$, $\out$ parameters $\vec{p}_O \subseteq
\vec{p}$,
private parameters $\vec{q}$, and parameter bindings $\vec{v}$ and
$\vec{w}$.  The set of all parameters is $\vec{p} = \vec{p}_I \cup
\vec{p}_O$. 
The $\inn$ and $\out$ parameters are mutually disjoint: $\vec{p}_I
\cap \vec{p}_O = \emptyset$.  A protocol's references (i.e.,
subprotocols, including message schemas) are denoted by the set $F$.

\begin{definition}\label{def:BSPLProtocol}
  A \textit{protocol} is a tuple
  $P = \langle n, \vec{x}, \vec{y}, \vec{p}, \vec{k}, \vec{q}, F
  \rangle$ where $n$ is a name.  $\vec{x}, \vec{y}, \vec{p}, \vec{q}$
  are as above, $F$ is a finite set of $f$ subprotocol
  references $F = \{ F_1, \ldots, F_f \}$, 
  such that $P$ includes each referenced sub protocol
  $F_i$'s roles, and key and non key parameters
  $(\forall i: 1 \leq i \leq f \Rightarrow F_i = \langle n_i,
  \vec{x}_i, \vec{p}_i, \vec{k}_i \rangle)$ where
  $\vec{x}_i \subseteq \vec{x} \cup \vec{y}$,
  $\vec{p}_i \subseteq \vec{p} \cup \vec{q}$,
  $\vec{k}_i = \vec{p}_i \cap \vec{k}$).   An atomic protocol
  with two roles and no 
  references is a message schema denoted as 
  $\ceil{s \mapsto r:m \; \vec{p}(\vec{k})}$.
\end{definition}

Later, protocol enactment is defined over a Universe of
Discourse (UoD) comprising roles and message schemas (for
convenience we modify the original BSPL definition
from a UoD comprising message \textit{references}).

\begin{definition}\label{def:PUoD}\cite[Def.~12]{Singh2012}
The \textit{UoD} of protocol $P$, 
$\text{UoD}(P) = \langle \pazocal{R}, \pazocal{M} \rangle$ 
consists of $P$’s roles and message schemas including the
message schemas of its referenced protocols recursively.
\end{definition}

\subsection{Commitments}

A \emph{commitment specification}
is a finite string according to the corresponding
representation in Table~\ref{table:syntax} over a message schema name set Base.

A commitment specification, $c(x, y, \textit{Cre}, \textit{Det}, \textit{Dis})$
from a debtor $x$ to a creditor $y$ is defined
over BSPL protocol message schema references (Base events)
used by an input protocol. Role names and time instants
are sets $\pazocal{R}$ and $\pazocal{T}$, respectively.
Operators $\sqcap$, $\sqcup$, and $\ominus$ 
are respectively conjunction, disjunction, and exception. 
In $E[l,r]$, $[l,r]$ is the time interval that $E[l,r]$
occurs within. We omit $l$ and $r$ when they are respectively $0$ and $\infty$.

\begin{table}[t]
\centering
\small
\caption{Cupid's grammar. Expr is create, detach, and discharge
conditions.}
\label{table:syntax}
\begin{tabular}{@{ }l@{ }l}\toprule
%

%
Event & $\lra$ Base $|$ LifeEvent\\
LifeEvent & $\lra$ created($\pazocal{R}$, $\pazocal{R}$, Expr, Expr, Expr) $|$ \\
  & \hspace{0.7cm}detached($\pazocal{R}$, $\pazocal{R}$, Expr, Expr, Expr) $|$\\
  & \hspace{0.7cm}discharged($\pazocal{R}$, $\pazocal{R}$, Expr, Expr, Expr) $|$ \\
  & \hspace{0.7cm}expired($\pazocal{R}$, $\pazocal{R}$, Expr, Expr, Expr) $|$ \\
  & \hspace{0.7cm}violated($\pazocal{R}$, $\pazocal{R}$, Expr, Expr, Expr)\\

Expr  &$\lra$  Event[Time, Time] $|$ Expr $\sqcap$\!  Expr\!  $|$ \!Expr $\sqcup$ \!Expr\!  $|$ \\
           & \hspace{0.7cm}Expr $\ominus$ Expr\\
Time &$\lra$ Event + $\pazocal{T}$ $|$ $\pazocal{T}$ \\
ComSpec & $\lra$ c($\pazocal{R}$, $\pazocal{R}$, Expr, Expr, Expr)\\
\bottomrule
\end{tabular}
\end{table}

\subsection{Commitment Operationalization}

Each commitment we wish to operationalize is rewritten into a
commitment alignment protocol for an input protocol.
A forwarding message schema has a unique name and $\out$ parameter, 
to avoid conflicts with other message schemas. 

Let $\pazocal{N}$ be the set of unique message forwarding schema names
disjoint from the message schema name set Base.
Unique forwarding message schema names
are obtained taking as input a message schema name from Base and a
role, and then outputting a forwarding message schema name,
using an assumed injective function
$V : \text{Base} \times \pazocal{R} \rightarrow \pazocal{N}$.
For example, $\mname{fwdSEShip} = V(\mname{ship}, E)$
is a unique name for forwarding \mname{ship} from the
shipper (S) to the escrow (E).
The inverse injective function $V^{-1}$ determines which message
is being forwarded. 
An assumed injective parameter name function 
$\textit{VID} : \pazocal{N} \rightarrow \pazocal{ID}$ from
forwarding message names to identifier parameter names ($\pazocal{ID}$)
produces a uniquely named $\out$ parameter for each forwarding message
schema (e.g., 
$\pname{fwdSEShipID} = \textit{VID}(\pname{fwdSEShip})$).

A forwarding message schema is included in a commitment operationalization protocol
when necessary to support commitment alignment.
For example, if $E$ is the commitment's create condition, $b$ the debtor
and $a$ the creditor. Then, the event formula $E$ must be aligned between
roles $a$ and $b$ such that if $a$ knows $E$ then $b$ can learn of $E$ via 
message forwarding. 

Each commitment $C$ is decomposed via rewrites 
into instructions of the form $\textit{al}(a, E, b)$ stating
a requirement for $E$ to be \emph{al}igned from $a$ to $b$.
If the formula $E$ is non atomic, then
it is further decomposed to eventually
atomic alignment instructions, $\textit{al}(a, m, b)$ on messages $m$.
Atomic message alignment instructions
are rewritten into the necessary message forwarding
schema in the alignment protocol that we are constructing.
We present the base case 
rewrites for alignment instructions operating on message schemas, then
non atomic event formulae and finally commitments.

The presented rewrite rules are for an input protocol
$P = \langle n, \vec{x}, \vec{y}, 
\vec{p}, \vec{k}, \vec{q}, F \rangle$ and its
Universe of Discourse $\langle \pazocal{R}, \pazocal{M} \rangle = \text{UoD}(P)$,
a commitment $C$ and the commitment alignment
protocol $P^{\textit{C}} = \langle n^C, \vec{x}^{C}, \vec{y}^{C}, 
\vec{p}^{C}, \vec{k}^{C}, \vec{q}^{C}, F^{C} \rangle$ being constructed.

Rule~\ref{eqRwDirectAtom} handles aligning an atomic message. 
It is conditional on: 
\begin{inparaenum}[(1)] 
\item An atomic message alignment instruction $\textit{al}(a, m, b)$.
\item The commitment alignment protocol $P^{C}$ being constructed.
\item A message schema in the input protocol $P$ where a role $s$ that
is distinct from $b$ instantiates the message $m$
via emission to a role $r$ distinct from $b$. 
\end{inparaenum}

The rewrite result is: \begin{inparaenum}[(1)]\setcounter{enumi}{3}
\item A new message schema labeled $m^{\textit{forw}}$ acting to
forward message schema $m$'s instances,
from the role $s$ to the role $b$.
\item The commitment alignment protocol referencing $m^{\textit{forw}}$.
\item The commitment alignment protocol including
the forwarding message schema's $\key$ parameters, 
\item $\inn$ and $\out$ parameters, and roles.
\end{inparaenum}

\begin{align*}
\inference
{
{\begin{array}{@{}c@{} r}
\; \; \; \; \; \; \; \;  \; \; \; \; \; \; \; \;  \; \;
\textit{al}(a, m, b), &  (1) \\
\; \; \; \; \; \; \; \;  \; \; \; \; \; \; \; \;  \; \;
P^{\textit{C}} = \langle n^{C}, \vec{x}^{C}, \vec{y}^{C}, 
\vec{p}^{C}, \vec{k}^{C}, \vec{q}^{C}, F^{C} \rangle, & (2) \\
\; \; \; \; \; \; \; \;  \; \; \; \; \; \; \; \; \; \;
\ceil{\textit{s} \mapsto r : m \; \vec{p}(\vec{k})} \in \pazocal{M}, \; \; s \neq b, r \neq b &  
\; \; \; \; \; \; \; \;  \; \; \; \;  \; \; \; \;  \; \; \; \;  \; \; \; \;
(3)
\end{array}}
}
{
\begin{array}{@{}c@{} r}
\ceil{\textit{s} \mapsto b : m^{\textit{forw}}\vec{p}^{\textit{forw}}(\vec{k}^{\textit{forw}})} & \; \; \; \; (4) \\
F^C = F^C \cup \{ m^{\textit{forw}} \}, & (5) \\
\vec{k}^{C} = \vec{k}^{C} \cup \vec{k}, \; \; & (6)  \\
p^{C}_{I} = p^{C}_{I} \cup p^{\textit{forw}}_{I}, \; \;
p^{C}_{O} = p^{C}_{O} \cup p^{\textit{forw}}_{O}, 
\vec{x}^{C} = \vec{x} \cup \vec{x}^{C} \; \; & (7)
\end{array}
}
\tag{R1}\label{eqRwDirectAtom}
\end{align*}

Where:

\begin{itemize}
\item The uniquely named forwarding message schema forwards $m$ to $b$: 
$m^{\textit{forw}} = V(m, b)$.
\item The forwarding message schema's parameters comprise: 
keys corresponding to $m$'s ($k^{\textit{forw}} = k$);
a unique $\out$ parameter to ensure that protocol enactment requires forwarding
($p^{\textit{forw}}_{O} = \{ \textit{VID}(m^{\textit{forw}}) \}$); and
$\inn$ parameters matching $m$'s parameters 
($p^{\textit{forw}}_{I} = p$), meaning that $m$ must be instantiated prior to
it being forwarded.
\end{itemize}

Instructions to align messages from $a$ to $b$ occurring within a time window
are reduced to atomic message alignment instructions.
The message that occurs as well as any start or deadline messages
are necessarily also aligned from $a$
to $b$ according to the rewrite Rule~\ref{eqRwSTDT} (omitting
cases for time windows without either a start time, a deadline, or
both).

\begin{align*}
\inference
{
\textit{al}(a, m[s \pm J, d \pm K], b)
}
{
\begin{array}{@{}c@{}}
\textit{al}(a, m, b) \; \; \textit{al}(a, s, b) \; \; \textit{al}(a, d, b)
\end{array}
}
\tag{R4}\label{eqRwSTDT}
\end{align*}

To give an example, the \emph{EscrowPurchase} commitment
from the merchant to the customer in Listing~\ref{cupid:thirdparty} is detached
when the customer pays the escrow within 
ten time points of a price quote. Hence
we have an alignment instruction
$\textit{al}(\textit{C}, \textit{payEscrow}[,\textit{quote} + 10], \textit{M})$
to ensure that when the creditor (customer) knows the detachment
so does the debtor (merchant). The alignment instruction is reduced to
$\textit{al}(\textit{C}, \textit{payEscrow}, \textit{M})$
and $\textit{al}(\textit{C}, \textit{quote}, \textit{M})$.

In the input protocol in Listing~\ref{list:input-thirdparty}
\mname{quote} is from the merchant to the customer,
guaranteeing alignment, and so $\textit{al}(\textit{C}, \textit{quote}, \textit{M})$ is not rewritten. However, \mname{payEscrow} is not received or sent by the merchant and hence
must be forwarded by the customer.
The instruction $\textit{al}(\textit{C}, \textit{payEscrow}, \textit{M})$ is
rewritten to a message schema as in Listing~\ref{BSPL:PartialShippingAl1}.

\begin{lstlisting}[
  label={BSPL:PartialShippingAl1},
  caption={A partial alignment protocol for the \emph{EscrowPurchase} commitment
  in Listing~\ref{cupid:thirdparty}},
  numbers=left,
  numbersep=1pt,
  xleftmargin=5pt,
  escapechar=|,
  numberstyle=\scriptsize
]
EscrowPurchaseAl{
$\msf{roles}$ C, M
$\msf{parameters}$ in oID key, in pID, out fwdCMPayEscrowID
C $\mapsto$ M:fwdCMPayEscrow[in oID, in pID, out fwdCMPayEscrowID] }
\end{lstlisting}

Both sides of a conjunct are aligned according to Rule~\ref{eqRwConj}.
Likewise both sides of a disjunct are aligned 
via Rule~\ref{eqRwDisj},
since we do not know which runtime messages will occur
\textit{a priori}.

Exceptions are handled by 
Rule~\ref{eqRwExc}. In order to guarantee that when a role $a$ knows $L \ominus R$ 
then so does $b$, we must ensure that if $a$ knows $L$ then
so can $b$ via messaging. Yet, if $b$ knows $R$
then it will never know $L \ominus R$ to be true, even if $a$ knows
$L$, believes $L \ominus R$ is true and so forwards $L$ to $b$. Thus
we align $L$ from $a$ to $b$ and $R$ from $b$ to $a$.

\noindent\begin{minipage}{.505\linewidth}
\begin{align*}
\inference
{
\textit{al}(a, L \sqcap R, b)
}
{
\begin{array}{@{}c@{}}
\textit{al}(a, L, b) \; \; \textit{al}(a, R, b)
\end{array}
}
\tag{R5}\label{eqRwConj}
\end{align*}
\end{minipage}%
\begin{minipage}{.51\linewidth}
\begin{align*}
\inference
{
\textit{al}(a, L \sqcup R, b)
}
{
\begin{array}{@{}c@{}}
\textit{al}(a, L, b) \; \; \textit{al}(a, R, b)
\end{array}
}
\tag{R6}\label{eqRwDisj}
\end{align*}
\end{minipage}

\begin{align*}
\inference
{
\textit{al}(a, L \ominus R, b)
}
{
\begin{array}{@{}c@{}}
\textit{al}(a, L, b) \; \; \textit{al}(b, R, a)
\end{array}
}
\tag{R7}\label{eqRwExc}
\end{align*}

For example, a shipment commitment from
the shipper to the merchant is discharged if: the item is shipped
within five time points of the shipment being requested, except 
if the shipment is reported as damaged within five time points of being received. 
Thus we have an alignment instruction from the shipper to the merchant:
$\textit{al}(S, \textit{ship}[,\textit{requestShip} + 5] 
\ominus \textit{reportDamage}[,\textit{ship} + 5], M)$.
Alignment holds when: if the shipper knows $\textit{ship}[,\textit{requestShip} + 5]$
then so does the merchant and if the merchant knows the exception
$\textit{reportDamage}[,\textit{ship} + 5]$ then so does the shipper. 
Hence the alignment instruction is rewritten to
$\textit{al}(S, \textit{ship}[\textit{requestShip} + 5], M)$
and $\textit{al}(M, \textit{reportDamage}, S)$.

Nested commitment lifecycle events
occur when the corresponding lifecycle event for the referenced commitment 
$c(x, y, \textit{Cre}, \textit{Det}, \textit{Dis})$ occurs. 
Hence, we rewrite nested lifecycle events to the conditions 
under which they occur according to Cupid's 
semantics with Rules~\ref{eqRwCreated} to \ref{eqRwViolated}.

\begin{align*}
\inference
{
\textit{al}(a, \textit{created}(x, y, \textit{Cre}, \textit{Det}, \textit{Dis}), b)
}
{
\begin{array}{@{}c@{}}
\textit{al}(a, \textit{Cre}, b)
\end{array}
}
\tag{R8}\label{eqRwCreated}
\end{align*}

\begin{align*}
\inference
{
\textit{al}(a, \textit{detached}(x, y, \textit{Cre}, \textit{Det}, \textit{Dis}), b)
}
{
\begin{array}{@{}c@{}}
\textit{al}(a, \textit{Cre} \sqcap \textit{Det}, b)
\end{array}
}
\tag{R9}\label{eqRwDetached}
\end{align*}

\begin{align*}
\inference
{
\textit{al}(a, \textit{discharged}(x, y, \textit{Cre}, \textit{Det}, \textit{Dis}), b)
}
{
\begin{array}{@{}c@{}}
\textit{al}(a, (\textit{Cre} \sqcap \textit{Dis}) \sqcup
(\textit{Det} \sqcap \textit{Dis}), b)
\end{array}
}
\tag{R10}\label{eqRwDischarged}
\end{align*}

\begin{align*}
\inference
{
\textit{al}(a, \textit{expired}(x, y, \textit{Cre}, \textit{Det}, \textit{Dis}), b)
}
{
\begin{array}{@{}c@{}}
\textit{al}(a, \textit{Cre} \ominus \textit{Det}, b)
\end{array}
}
\tag{R11}\label{eqRwExpired}
\end{align*}

\begin{align*}
\inference
{
\textit{al}(a, \textit{violated}(x, y, \textit{Cre}, \textit{Det}, \textit{Dis}), b)
}
{
\begin{array}{@{}c@{}}
\textit{al}(a, (\textit{Cre} \sqcap \textit{Det}) \ominus
\textit{Disch}, b)
\end{array}
}
\tag{R12}\label{eqRwViolated}
\end{align*}

The final rewrite is for commitments. A commitment is aligned
when: if the creditor knows it is created, detached, or violated,
then the debtor respectively knows it is created, detached, or violated;
and if the debtor knows it is discharged or expired, then the
creditor knows it is respectively discharged or expired.
Owing to this asymmetry, we rewrite a commitment with
Rule~\ref{eqRwCommitment} to alignment instructions for each lifecycle event.

\begin{align*}
\inference
{
c(x, y, \textit{Cre}, \textit{Det}, \textit{Dis})
}
{
\begin{array}{@{}c@{}}
\textit{al}(c, \textit{created}(x, y, \textit{Cre}, \textit{Det}, \textit{Dis}),
d) \\
\textit{al}(c, \textit{detached}(x, y, \textit{Cre}, \textit{Det}, \textit{Dis}),
d) \\
\textit{al}(c, \textit{violated}(x, y, \textit{Cre}, \textit{Det}, \textit{Dis}),
d) \\
\textit{al}(d, \textit{discharged}(x, y, \textit{Cre}, \textit{Det}, \textit{Dis}),
c) \\
\textit{al}(d, \textit{expired}(x, y, \textit{Cre}, \textit{Det}, \textit{Dis}),
c)
\end{array}
}
\tag{R13}\label{eqRwCommitment}
\end{align*}

We now define a commitment alignment protocol.

\begin{definition} A protocol $P^{C}$ is a \textit{commitment
alignment protocol} for a commitment $C$ and an input protocol $P$
iff all possible applications of \ref{eqRwDirectAtom} to \ref{eqRwCommitment} are
made to $C$, an empty version of $P^{C}$ and $P$'s
UoD $\langle \pazocal{R}, \pazocal{M} \rangle = \text{UoD}(P)$.
\end{definition}

Each rule is a monotonic reduction on finite formulae. Hence:

\begin{lemma} There exists a commitment operationalization
protocol $P^{C}$ for each commitment $C$ and input protocol 
$P$.
\end{lemma}

An operationalization protocol is composed
from the input protocol and commitment alignment protocols
(omitting empty and redundant subprotocols).

\begin{definition}\label{def:operationalization-protocol} 
Let $P = \langle n, \vec{x}, \vec{y}, 
\vec{p}, \vec{k}, \vec{q}, F \rangle$ be an input protocol 
and $\mathbb{C}$ be a set of commitments
defined over the message schema names and roles in $P$'s Universe of
Discourse $\langle \pazocal{R}, \pazocal{M} \rangle = \text{UoD}(P)$.
Let each commitment $C \in \mathbb{C}$ have an alignment protocol
$P^{C} = \langle n^{C}, \vec{x}^{C}, \vec{y}^{C}, 
\vec{p}^{C}, \vec{k}^{C}, \vec{q}^{C}, F^{C}
\rangle$ for $P$ that includes at least two roles 
($|\vec{x}^{c}| \geq 2$). 
$P^{\mathbb{C}} = \langle n^{\mathbb{C}}, \vec{x}^{\mathbb{C}}, \vec{y}^{\mathbb{C}}, 
\vec{p}^{\mathbb{C}}, \vec{k}^{\mathbb{C}}, \vec{q}^{\mathbb{C}}, F^{\mathbb{C}}
\rangle$ is an operationalization protocol for $\mathbb{C}$ and $P$ iff:
\begin{itemize}[]
\item $P^{\mathbb{C}}$ references the input protocol and all commitment operationalization
protocols: $F^{\mathbb{C}} = \{ n \} \cup \bigcup_{C \in \mathbb{C}} \{ n^{C} \}$.
\item $P^{\mathbb{C}}$'s roles and key parameters match the input protocol's: 
$\vec{x}^{\mathbb{C}} = \vec{x}$
and $\vec{k}^{\mathbb{C}} = \vec{k}$.
\item $P$'s $\out$ parameters comprise the input protocol's and each commitment
alignment protocol's $\out$ parameters: 
$\vec{p}^{\mathbb{C}}_O = \{ \vec{p}_{O} \} \cup \bigcup_{C \in \mathbb{C}} \{ \vec{p}_{O}^{C} \}$.
\end{itemize}
\end{definition}

\subsection{Semantics}

In BSPL, each message instance
$m[s,r,\vec{p}, \vec{v}]$ denotes a sending role $s$, a recipient
$r$, a parameter vector $\vec{p}$ with a corresponding parameter 
binding value vector $\vec{v}$.
A role's history denotes its sent and received messages in sequence.
A history vector comprises each role's history
where every received message must have been sent.

\begin{definition}\label{def:BSPLHistory}\cite[Def.~5]{Singh2012} 
A history of a role $\rho$, $H_{\rho}$, 
is given by a sequence of zero or more message 
instances $m_{1} \circ  m_{2} \circ ...$. Each $m_{i}$ 
is of the form $m[s,r,\vec{p}, \vec{v}]$ where 
$\rho = s$ or $\rho = r$, and $\circ$ means sequencing.
\end{definition}

\begin{definition}\label{def:BSPLHistoryVector}\cite[Def.~7]{Singh2012} 
We define a \textit{history vector} for
a UoD $\pazocal{R}, \pazocal{M}$, as $[H^{1}, ..., H^{|\pazocal{R}|}]$,
such that $\forall s, r:  1 \leq s, r \leq |\pazocal{R}|$ : $H^{s}$ is a
history s.t. $\forall m[s,r,\vec{p}, \vec{v}] \in
H^{r} : m \in \pazocal{M}$ and $m[s,r,\vec{p}, \vec{v}] \in  H^{s}$.
\end{definition}

A history vector is viable
if and only if sent and received messages 
bind values to parameters specified in each
corresponding message schema, respecting values already determined
by keys via known messages (for brevity, we omit Singh's
\shortcite[Def.~8]{Singh2012} definition). 
The set of all viable history vectors for a UoD (e.g.,
a protocol's roles and message schemas) is its
Universe of Enactments. 

\begin{definition}\label{def:UoE}
Given a UoD $\langle \pazocal{R},\pazocal{M} \rangle$, the 
\textit{Universe of Enactments} (UoE) for that UoD,
$\pazocal{U}_{\pazocal{R},\pazocal{M}}$, is the set of viable history 
vectors \shortcite[Definition~8]{Singh2012}, 
each of which has exactly $|\pazocal{R}|$ dimensions and each 
of whose messages instantiates a schema in $\pazocal{M}$.
\end{definition}

In Cupid \cite{Chopra2015}, an agent's \textit{model} maps from
event schemas to event instances, representing the agent's local view
of event occurrences. Here, we are dealing with messages and hence a model is simply a role's
history albeit substituting each
forwarding message with the message it forwards
and timestamping each message with the time it became known
(via emission, reception, or notification).

\begin{definition}\label{def:K} Let $P$ be an input protocol
and let Base be the message schema names for the message schemas in $P$'s UoD.
Let $P^{\mathbb{C}}$ be an operationalization protocol for $P$. 
A \textit{model} for a role $a$'s history $H$ and $P^{\mathbb{C}}$
is a history $M$, where each message is in the model 
$m^{M}_{i}[s^{M}_{i}, r^{M}_{i}, \vec{p}^{M}, \vec{v}^{M}] \in M$
iff there is a corresponding original message
$m^{H}_{i}[s^{H}_{i}, r^{H}_{i}, \vec{p}^{H}, \vec{v}^{H}] \in H$ meeting
(a) \textbf{or} (b), \textbf{and} (c):
\begin{enumerate}
\renewcommand{\theenumi}{\alph{enumi}}
\renewcommand{\labelenumi}{(\theenumi)}
\item The corresponding original message in $H$
is a non forwarding message $m^{H}_{i} \in \text{Base}$
and the names match: $m^{M}_{i} = m^{H}_{i}$.
\item The corresponding original message $m^{H}_{i}$ in $H$ is a forwarding message
and the message $m^{M}_{i}$ in the model takes the name of the message
being forwarded: $m^{M}_{i} = V^{-1}(m^{H}_{i}, r^{H}_{i})$.
\item The message $m^{M}_{i}$ contains all of the original message $m^{H}_{i}$
in $H$'s non forwarding ID parameters and parameter 
bindings with an additional timestamp parameter and parameter binding:
if $\exists \vec{p}^{H}_{j} \in \vec{p}^{H} = \textit{VID}^{-1}(m^{H}_{i})$
then $\vec{p}^{M}_{i} = (\vec{p}^{H}_{i} \backslash
\vec{p}^{H}_{j}) \cup \{ \text{time} \}$ and 
$\vec{v}^{M}_{i} = (\vec{v}^{H}_{i} \backslash
\vec{v}^{H}_{j}) \cup \{ \text{t} \}$, otherwise
$\vec{p}^{M}_{i} = \vec{p}^{H}_{j} \cup \{ \text{time} \}$ 
and $\vec{v}^{M}_{i} = \vec{v}^{H}_{j} \cup \{ \text{t} \}$, where $t$
is a timestamp.
\end{enumerate}
\end{definition}

We adopt Cupid's commitment semantics.  
For brevity, we only define the set of instances for event formula $E$ (e.g., a lifecycle event) entailed by a model $M$: $\llbracket E \rrbracket_M$.  If $E$ is
a non atomic event formula or a lifecycle event, then the result
is a database operation on the messages that cause $E$ to occur.
For example, the set of all ship instances 
is denoted as $\llbracket \textit{ship} \rrbracket$.
Moreover, the set of tuples modeled by the formulae for shipment occurring within
ten time points, $\textit{ship}[0, 10]$, is returned
by selecting all shipment events that occur between zero
and ten time points ($\llbracket \textit{ship}[0, 10] \rrbracket = 
\sigma_{0 \leq t < 10}(\llbracket \textit{ship} \rrbracket)$).
A database operation is inductively defined in Cupid for queries
corresponding to each event formula type.

\begin{definition}\label{def:CupidQuery} Let $M$ be a
model and $E$ an event formula. $\llbracket E \rrbracket_M$ is the
set of $E$'s instances (a set of tuples
combining stored events) 
returned from the query for $E$ on $M$ 
according to \cite[$D_{1} - D_{20}$]{Chopra2015}.
\end{definition}

\section{Properties}

An operationalization protocol retains an input protocol's message ordering.

\begin{lemma}\label{lemma:messageRetaining} Let $P$ be an input protocol,
and $P^{\mathbb{C}}$ be a commitment operationalization protocol.
If there is a history vector $H$ in $\textit{UoD}(P)$'s UoE
then there exists a history vector $H^{\mathbb{C}}$ in $\textit{UoD}(P^{\mathbb{C}})$'s 
UoE where for each history $H^{i \mathbb{C}}$ in $H^{\mathbb{C}}$ the messages instantiating
schemas in $\pazocal{M}$ are included in the same order
of the corresponding history $H^{i}$ in $H$.
\end{lemma}
\begin{proofsketch} We include 
messages from the input protocol's history to the operationalization
protocol's corresponding history, if they instantiate a schema
in the input protocol. We can always include received messages 
(\shortcite[Def.~6]{Singh2012}),
emitted messages can be included since they do not violate bindings
and $\inn$ parameters are necessarily known by binding parameters
in the operationalization protocol's message schemas.
\end{proofsketch}

Singh \shortcite{Singh2012} formalizes liveness and safety for
BSPL, and gives verification techniques. 
A protocol is \emph{safe} iff each history vector in the UoE
preserves uniqueness for each binding.
A protocol is \emph{live} iff
any enactment can progress to completion such that all parameters are
bound. 

\begin{theorem}
Let $P^{\mathbb{C}}$ be a commitment operationalization
protocol for an input protocol $P$.
If $P$ is safe then  $P^{\mathbb{C}}$ is safe.
If $P$ is live then $P^{\mathbb{C}}$ is live.
\end{theorem}
\begin{proofsketch} Safety: Assume $P^{\mathbb{C}}$ is unsafe.
Since $P^{\mathbb{C}}$  is an operationalization protocol, 
by Rule~\ref{eqRwDirectAtom}~(7) it does not introduce $\out$
parameters used by another message schema, hence by applying
Lemma~\ref{lemma:messageRetaining} $P$ must be unsafe as well.
Liveness: As for safety, relying on $P^{\mathbb{C}}$ 
not introducing $\out$ parameters used by $P$.
\end{proofsketch}

C\&S define alignment for active commitments.
Definition~\ref{def:Alignment} generalizes it to all states in Cupid's
commitment lifecycle.  Specifically, if a creditor infers created, detached or violated of a
commitment (thereby strengthening the expectation) from its history,
then the debtor must as well (i.e., know what is expected of them).
Conversely, if a debtor infers discharge or expired (hence weakening
the expectation), the creditor must as well.

\begin{definition}\label{def:Alignment} 
Let $M^{x}$ and $M^{y}$ be models.  
  The history vector $H$ is aligned with respect
  to $\textit{c}(x, y, C, D, U)$ iff:
\begin{align*}
i\in \llbracket \textit{created}(x,y,C,D,U)\rrbracket_{M^{y}} \mkern-4mu \Rightarrow &
i \in  \llbracket \textit{created}(x,y,C,D,U)\rrbracket_{M^{x}} \\
i \in \llbracket \textit{detached}(x,y,C,D,U)\rrbracket_{M^{y}} \mkern-4mu \Rightarrow &  i \in \llbracket \textit{detached}(x,y,C,D,U)\rrbracket_{M^{x}} \\
i\in \llbracket \textit{violated}(x,y,C,D,U)\rrbracket_{M^{y}} \mkern-4mu \Rightarrow &
i \in \llbracket \textit{violated}(x,y,C,D,U)\rrbracket_{M^{x}} \\
\mkern-4mu i \mkern-1mu \in \mkern-1mu \mkern-1mu \llbracket \textit{discharged}(x,y,C,D,U))\rrbracket_{M^{x}} \mkern-4mu \Rightarrow & \mkern-2mu
i \mkern-1mu \in \mkern-1mu \llbracket \textit{discharged}(x,y,C,D,U)\rrbracket_{M^{y}} & \\
i \in \llbracket \textit{expired}(x,y,C,D,U)\rrbracket_{M^{x}} \mkern-4mu \Rightarrow &
i \in \llbracket \textit{expired}(x,y,C,D,U)\rrbracket_{M^{y}}
\end{align*}
\end{definition}

The idea that messages should happen 
in some time interval relies on a global clock.
However, there is the potential for misalignment
due to message delays and local clock skews, rather than
which messages are emitted and received. 
Such problems are avoided by making the time intervals an order of magnitude
larger than the maximum clock skew and message delays
\cite{Cranefield2005}.

We assume that for a commitment $c(x, y, \textit{Cre}, \textit{Det}, \textit{Dis})$ 
being operationalized with respect to a history vector $H$ if when 
$x$ or $y$ knows a message $m$
and the counter-party receives $m$ 
they will do so at a time to make the same
inferences over $\textit{Cre}$, $\textit{Det}$, and $\textit{Dis}$. 
Under this assumption, an operationalization protocol
is sufficient to support alignment.

Theorem~\ref{th:alignment} states that a commitment operationalization
protocol always makes it possible to rectify alignment via
messaging.

\begin{theorem}\label{th:alignment}
  Let $P^{\mathbb{C}}$ be a
  protocol that operationalizes a set of commitments $\mathbb{C}$
  such that $c(x, y, \textit{Cre}, \textit{Det}, \textit{Dis})  \in \mathbb{C}$.
  If history vector $H \in \pazocal{U}_{\pazocal{R},\pazocal{M}}$
  is in $P^{\mathbb{C}}$'s UoE
  then there exists a longer $H^{\prime \prime}$ in $P^{\mathbb{C}}$'s
  UoE that is aligned with respect to $c(x, y, \textit{Cre}, \textit{Det}, \textit{Dis})$.
\end{theorem}

\begin{proofsketch}
  If $H$ is misaligned. 
  Definition~\ref{def:Alignment} and Cupid's semantics for lifecycle
  events \cite[$D_{15}$ to $D_{19}$]{Chopra2015} imply role $s$'s model entails $E$.
  Base case: $E = m$. By \ref{eqRwDirectAtom} extend $H$ to a
  history vector $H^{\prime \prime}$ in $P^{\mathbb{C}}$'s UoE
  (Definition~\ref{def:UoE}) by inserting a notification $m_i$
  from $m$'s sender to $r$ in their respective histories.
  Inductive hypothesis: assume 
  there exists a history $H^{\prime}$ in $P^{\mathbb{C}}$'s UoE extending $H$
  s.t. if $E = F \sqcap G$ or $E = F \sqcup G$ then $r$ knows $F$ and $G$, 
  or for $E = F \ominus G$ $r$ knows $F$ and $s$ knows $G$. Inductive step:
  Extend $H^{\prime}$ to  $H^{\prime \prime}$ with an $m_i$
  via  rules \ref{eqRwDirectAtom} to \ref{eqRwViolated}.
  By the time assumption and Cupid's semantic definitions \cite[Def. 16 to
  Def. 19]{Chopra2015} $s$ and $r$ are aligned.
\end{proofsketch}

\section{Conclusions}

Tosca addresses challenges of decentralized commitment enactment.
Given a set of commitments defined over a protocol, it enables
automatically synthesizing a new protocol that supports alignment, a
form of commitment-level interoperability. Furthermore, the synthesized
protocol preserves liveness and safety, both of which are also
necessary for interoperation.  The new protocol may be thought of as a
\emph{fleshing out} of the input protocol.
Tosca brings together
several advances---in the specification of protocols, commitments, and
ideas about interoperability---toward supporting decentralization.

Tosca establishes a separation of concerns between
commitments and protocols.  Protocols are specified in BSPL
whereas commitments are specified in Cupid---languages developed
independently of each other.  Notably, in Cupid, commitments are
defined over a database schema, not a protocol.  We use
the fact that BSPL specifications are interpreted over databases in
layering Cupid specifications over BSPL specifications. 

Decentralization is a theme of growing interest, (e.g., for norm
compliance \cite{Baldoni2015a} and monitoring \cite{Bulling2013}).
Tosca's architecture is distinct from shared memory (environment) 
approaches (e.g., \cite{Omicini2008}).  
Such approaches would benefit from Tosca in that
they would also need a clear specification of interactions, both in
terms of meanings and protocols, even if alignment itself would be
trivial because of shared memory.

Other works treat commitments
\cite{Chesani2013} and protocols \cite{Yadav2015} separately without
studying their relationship.  G\"{u}nay {\etal}
\cite{Gunay2015} generate commitment specifications
from requirements.  We understand commitment specifications as
requirements and synthesize operational protocol specifications.

Analogously Searle \shortcite[pp.\ 26--27]{Searle1995} demarcates between
constitutive rules, which make social actions possible
by ascribing institutional (social) facts, and norms,
which prescribe institutional facts. This separation
is adopted for commitment protocols overlaying 
constitutive rules \cite{Baldoni2013a}. Moreover,
norms are both defined over institutional
facts and interpreted at different
levels of abstraction using constitutive rules within agents
\cite{Criado2013a} and legal institutions \cite{King2016b,King2017}.
We separate richer concerns: commitments, focusing on relational information,
overlaying operational protocols, focusing on information causality.

Future directions should address limitations and further applications.
Tosca maintains agent autonomy, making alignment possible but
not regimented and hence limited by the extent to which autonomous agents
communicate message notifications. 
We demonstrated Tosca on a business domain but it could just as well be applied to requirements in other domains that involve interaction. In particular, healthcare and
government (e.g., national) contracts, studied in connection with
Cupid are prime candidates for Tosca.
Since Tosca provides support for messaging requirements via protocols,
the extent to which it can be applied to agent development
should be investigated (e.g., using
communication abstractions supported in agent programming frameworks
such as \cite{Boissier2013}).

\section*{Acknowledgments}
We would like to thank the anonymous reviewers for their helpful comments.
King, G\"{u}nay and Chopra were supported by the EPSRC grant EP/N027965/1 (Turtles).
Singh thanks
the US Department of Defense for partial support under the Science of
Security Lablet.

\bibliographystyle{named}

\begin{thebibliography}{}

\bibitem[\protect\citeauthoryear{Baldoni \bgroup \em et al.\egroup
  }{2013}]{Baldoni2013a}
Matteo Baldoni, Cristina Baroglio, Elisa Marengo, and Viviana Patti.
\newblock {Constitutive and Regulative Specifications of Commitment Protocols:
  a Decoupled Approach}.
\newblock {\em ACM Transactions on Intelligent Systems and Technology (TIST)},
  4(2), 2013.

\bibitem[\protect\citeauthoryear{Baldoni \bgroup \em et al.\egroup
  }{2015}]{Baldoni2015a}
Matteo Baldoni, Cristina Baroglio, Amit~K. Chopra, and Munindar~P. Singh.
\newblock {Composing and Verifying Commitment-Based Multiagent Protocols}.
\newblock In {\em Proceedings of the 24th International Joint Conference on
  Artificial Intelligence (IJCAI)}, pages 10--17. AAAI Press, 2015.

\bibitem[\protect\citeauthoryear{Boissier \bgroup \em et al.\egroup
  }{2013}]{Boissier2013}
Olivier Boissier, Rafael~H. Bordini, Jomi~F. H{\"{u}}bner, Alessandro Ricci,
  and Andrea Santi.
\newblock {Multi-agent oriented programming with JaCaMo}.
\newblock {\em Science of Computer Programming}, 78(6):747--761, 2013.

\bibitem[\protect\citeauthoryear{Bulling \bgroup \em et al.\egroup
  }{2013}]{Bulling2013}
Nils Bulling, Mehdi Dastani, and Max Knobbout.
\newblock {Monitoring norm violations in multi-agent systems}.
\newblock In {\em Proceedings of the 2013 International Conference on
  Autonomous Agents and Multi-agent Systems (AAMAS 2013)}, pages 491--498.
  International Foundation for Autonomous Agents and Multiagent Systems, 2013.

\bibitem[\protect\citeauthoryear{Chesani \bgroup \em et al.\egroup
  }{2013}]{Chesani2013}
Federico Chesani, Paola Mello, Marco Montali, and Paolo Torroni.
\newblock {Representing and monitoring social commitments using the event
  calculus}.
\newblock {\em Autonomous Agents and Multiagent Systems}, 27(1):85--130, 2013.

\bibitem[\protect\citeauthoryear{Chopra and Singh}{2008}]{Chopra2008}
Amit~K. Chopra and Munindar~P. Singh.
\newblock {Constitutive interoperability}.
\newblock In {\em Proceedings of the 7th International Joint Conference on
  Autonomous Agents and Multiagent Systems - Volume 2}, pages 797--804, 2008.

\bibitem[\protect\citeauthoryear{Chopra and Singh}{2009}]{Chopra2009}
AK~Chopra and MP~Singh.
\newblock {Multiagent commitment alignment}.
\newblock In {\em Proceedings of The 8th International Conference on Autonomous
  Agents and Multiagent Systems - Volume 2}, pages 937--944, 2009.

\bibitem[\protect\citeauthoryear{Chopra and Singh}{2015a}]{Chopra2015}
Amit~K. Chopra and Munindar~P. Singh.
\newblock {Cupid: Commitments in Relational Algebra}.
\newblock In {\em Proceedings of the Twenty-Ninth AAAI Conference on Artificial
  Intelligence}, pages 2052--2059, 2015.

\bibitem[\protect\citeauthoryear{Chopra and Singh}{2015b}]{Chopra2015a}
Amit~K. Chopra and Munindar~P. Singh.
\newblock {Generalized Commitment Alignment}.
\newblock In {\em Proceedings of the 14th International Conference on
  Autonomous Agents and Multiagent Systems}, pages 453--461, 2015.

\bibitem[\protect\citeauthoryear{Cranefield}{2005}]{Cranefield2005}
Stephen Cranefield.
\newblock {A Rule Language for Modelling and Monitoring Social Expectations in
  Multi-Agent Systems}.
\newblock In {\em Coordination, Organization, Institutions and Norms in
  Multi-Agent Systems. Lecture Notes in Computer Science.}, volume 3913, pages
  246--258. Springer, 2005.

\bibitem[\protect\citeauthoryear{Criado \bgroup \em et al.\egroup
  }{2013}]{Criado2013a}
N~Criado, E.~Argente, P.~Noriega, and V.~Botti.
\newblock {Reasoning about constitutive norms in BDI agents}.
\newblock {\em Logic Journal of the IGPL}, 22(1):66--93, 2013.

\bibitem[\protect\citeauthoryear{G{\"{u}}nay \bgroup \em et al.\egroup
  }{2015}]{Gunay2015}
Ak{\i}n G\"{u}nay , Michael Winikoff, and P{\i}nar Yolum.
\newblock {Dynamically generated commitment protocols in open systems}.
\newblock {\em Autonomous Agents and Multi-Agent Systems}, 29:192--229, 2015.

\bibitem[\protect\citeauthoryear{Huget and Odell}{2004}]{Huget2004}
Marc-Philippe Huget and James Odell.
\newblock {Representing agent interaction protocols with agent UML}.
\newblock In {\em Proceedings of the 1st International Workshop on
  Agent-Oriented Software Engineering (AOSE 2000), Lecture Notes in Computer
  Science}, volume 3382, pages 16 -- 30, 2004.

\bibitem[\protect\citeauthoryear{King \bgroup \em et al.\egroup
  }{2017}]{King2017}
Thomas~C. King, Marina {De Vos}, Virginia Dignum, Catholijn~M. Jonker, Tingting
  Li, Julian Padget, and M.~Birna van Riemsdijk.
\newblock {Automated multi-level governance compliance checking}.
\newblock {\em Autonomous Agents and Multi-Agent Systems}, 2017.

\bibitem[\protect\citeauthoryear{King}{2016}]{King2016b}
Thomas~C. King.
\newblock {\em {Governing Governance: A Formal Framework for Analysing
  Institutional Design and Enactment Governance}}.
\newblock PhD thesis, Delft University of Technology, 2016.

\bibitem[\protect\citeauthoryear{Miller and Mcginnis}{2007}]{Miller2007}
Tim Miller and Jarred Mcginnis.
\newblock {Amongst First-Class Protocols}.
\newblock In {\em Proceedings of the 8th International Workshop on Engineering
  Societies in the Agents World (ESAW 2007). Lecture Notes in Computer
  Science.}, volume 1957, pages 208 -- 223, 2007.

\bibitem[\protect\citeauthoryear{Omicini \bgroup \em et al.\egroup
  }{2008}]{Omicini2008}
Andrea Omicini, Alessandro Ricci, and Mirko Viroli.
\newblock {Artifacts in the A{\&}A meta-model for multi-agent systems}.
\newblock {\em Autonomous Agents and Multi-Agent Systems}, 17(3):432--456,
  2008.

\bibitem[\protect\citeauthoryear{Pitt \bgroup \em et al.\egroup
  }{2001}]{Pitt2001}
Jeremy Pitt, Lloyd Kamara, and Alexander Artikis.
\newblock {Interaction patterns and observable commitments in a multi-agent
  trading scenario}.
\newblock In {\em Proceedings of the 5th International Conference on Autonomous
  Agents}, pages 481--488, 2001.

\bibitem[\protect\citeauthoryear{Searle}{1995}]{Searle1995}
John~R. Searle.
\newblock {\em {The Construction of Social Reality}}.
\newblock The Free Press, New York, 1995.

\bibitem[\protect\citeauthoryear{Singh}{1999}]{Singh1999}
MP~Singh.
\newblock {An ontology for commitments in multiagent systems: Toward a
  unification of normative concepts}.
\newblock {\em Artificial Intelligence and Law}, 7:97--113, 1999.

\bibitem[\protect\citeauthoryear{Singh}{2011}]{Singh2011}
Munindar~P. Singh.
\newblock {Information-driven interaction-oriented programming: BSPL, the
  blindingly simple protocol language}.
\newblock In {\em Proceedings of the 10th International Conference on
  Autonomous Agents and Multiagent Systems - Volume 2}, pages 491--498, 2011.

\bibitem[\protect\citeauthoryear{Singh}{2012}]{Singh2012}
Munindar Singh.
\newblock {Semantics and Verification of Information-Based Protocols}.
\newblock In {\em Proceedings of the 11th International Conference on
  Autonomous Agents and Multiagent Systems - Volume 2}, pages 1149--1156, 2012.

\bibitem[\protect\citeauthoryear{Yadav \bgroup \em et al.\egroup
  }{2015}]{Yadav2015}
Nitin Yadav, Michael Winikoff, and Lin Padgham.
\newblock {HAPN: Hierarchical Agent Protocol Notation}.
\newblock In {\em Proceedings of the International Workshop on Coordination,
  Organisation, Institutions and Norms in Multi-Agent Systems (COIN@IJCAI)},
  2015.

\bibitem[\protect\citeauthoryear{Yolum and Singh}{2002}]{Yolum2002}
P{\i}nar Yolum and Munindar~P. Singh.
\newblock {Flexible Protocol Specification and Execution: Applying Event
  Calculus Planning using Commitments}.
\newblock In {\em Proceedings of the first International Joint Conference on
  Autonomous Agents and Multiagent Systems: part 2}, pages 527--534, Bologna,
  Italy, 2002.

\end{thebibliography}

\end{document}